\documentclass[10pt,twocolumn,letterpaper]{article}
\usepackage{times}
\usepackage{epsfig}
\usepackage{graphicx}
\usepackage{amsmath}
\usepackage{amssymb}
\usepackage[dvipsnames]{xcolor}
\usepackage{booktabs}
\usepackage{epigraph}

\usepackage{bm}
\usepackage{multirow}
\makeatletter

\newcommand{\Rmnum}[1]{\expandafter\@slowromancap\romannumeral #1@}
\makeatother
\usepackage{xcolor}
\usepackage{utfsym}
\usepackage{soul}
\usepackage[accsupp]{axessibility}

\definecolor{color_yellow}{RGB}{255,249,231}
\definecolor{color_pink}{RGB}{255,235,235}

\usepackage[pagenumbers]{cvpr} 

\definecolor{cvprblue}{rgb}{0.21,0.49,0.74}
\usepackage[pagebackref,breaklinks,colorlinks,citecolor=cvprblue]{hyperref}

\title{Correlation-Decoupled Knowledge Distillation for \\ Multimodal Sentiment Analysis with Incomplete Modalities}

\author{Mingcheng Li$^{1,2}$$\,^{\usym{1F396}}$ $\quad$
        Dingkang Yang$^{1,2}$$\,^{\usym{1F396}}$$\quad$
        Xiao Zhao$^{1,2}\quad$
        Shuaibing Wang$^{1,2}\quad$ 
        Yan Wang$^{1}\quad$ \\
        Kun Yang$^{1}\quad$ 
        Mingyang Sun$^{1,2}\quad$ 
        Dongliang Kou$^{1,2}\quad$ 
        Ziyun Qian$^{1,2}\quad$ 
        Lihua Zhang$^{1,2,3,4}$\footnotemark[4] \\ 
        \small$^1$Academy for Engineering and Technology, Fudan University$\,$ 
        \small$^2$Cognition and Intelligent Technology Laboratory (CIT Lab)\\
        \small$^3$Jilin Provincial Key Laboratory of
Intelligence Science and Engineering, Changchun, China\\
\small$^4$Engineering Research Center of AI and Robotics, Ministry of Education, Shanghai, China\\
{\tt\small mingchengli21@m.fudan.edu.cn, dkyang20@fudan.edu.cn}
}

\begin{document}
\maketitle
\renewcommand{\thefootnote}{\fnsymbol{footnote}}
\footnotetext[4]{Corresponding author. $^{\usym{1F396}}$Equal contribution.}

\begin{abstract}
Multimodal sentiment analysis (MSA) aims to understand human sentiment through multimodal data. Most MSA efforts are based on the assumption of modality completeness. However, in real-world applications, some practical factors cause uncertain modality missingness, which drastically degrades the model's performance. 
To this end, we propose a Correlation-decoupled Knowledge Distillation (CorrKD) framework for the MSA task under uncertain missing modalities.  
Specifically, we present a sample-level contrastive distillation mechanism that transfers comprehensive knowledge containing cross-sample correlations to reconstruct missing semantics. 
Moreover, a category-guided prototype distillation mechanism is introduced to capture cross-category correlations using category prototypes to align feature distributions and generate favorable joint representations. 
Eventually, we design a response-disentangled consistency distillation strategy to optimize the sentiment decision boundaries of the student network through response disentanglement and mutual information maximization. 
Comprehensive experiments on three datasets indicate that our framework can achieve favorable improvements compared with several baselines.

\end{abstract}

\section{Introduction}
\label{sec:intro}
\setlength{\epigraphwidth}{0.45\textwidth} 
\epigraph{\emph{``Correlations serve as the beacon through the fog of the missingness.''}}{\footnotesize--\emph{Lee \& Dicken}}
\vspace{-6pt}
\begin{figure}[t]
    \centering
    \includegraphics[width=\linewidth]{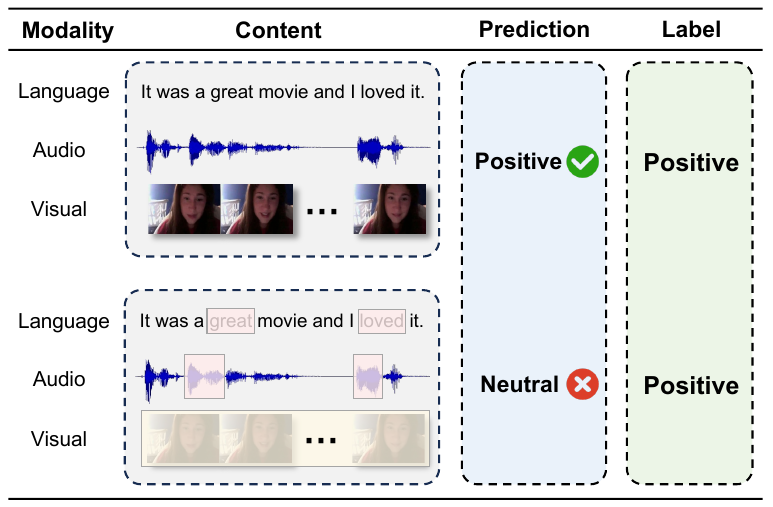}
    \caption{ Traditional model outputs correct prediction when inputting the sample with complete modalities, but incorrectly predicts the sample with missing modalities. We define two missing modality cases: (i) intra-modality missingness (\emph{i.e.}, the \textcolor{Magenta}{pink} areas) and (ii) inter-modality missingness (\emph{i.e.}, the \textcolor{Goldenrod}{yellow} area).}
    \label{fig:example}
\end{figure}

Multimodal sentiment analysis (MSA) has attracted wide attention in recent years. Different from the traditional unimodal-based emotion recognition task~\cite{du2021learning}, MSA understands and recognizes human emotions through multiple modalities, including language, audio, and visual \cite{morency2011towards}. 
Previous studies have shown that combining complementary information among different modalities facilitates the generation of more valuable joint multimodal representations \cite{springstein2021quti,shraga2020web}.
Under the deep learning paradigm~\cite{yang2023how2comm,yang2023spatio,yang2023what2comm,wang2023cpr,wang2021tsa,chen2024miss,kuang2023towards},
numerous studies assuming the availability of all modalities during both training and inference stages~\cite{hazarika2020misa, yu2021learning, yang2022disentangled, yang2022learning, yang2022emotion, li2023decoupled,yang2024robust,yang2023context,yang2023target,yang2024SuCi,yang2024MCIS,yang2022contextual,lei2023text}.
Nevertheless, this assumption often fails to align with real-world scenarios, where factors such as background noise, sensor constraints, and privacy concerns may lead to uncertain modality missingness issues.
Modality missingness can significantly impair the effectiveness of well-trained models based on complete modalities.
For instance, as shown in \Cref{fig:example}, the entire visual modality is missing, and some frame-level features in the language and audio modalities are missing, leading to an incorrect sentiment prediction.

In recent years, many works \cite{lian2023gcnet, wang2023distribution, pham2019found, wang2020transmodality, zeng2022tag,    liu2024modality, li2023towards,li2024unified} attempt to address the problem of missing modalities in MSA. 
As a typical example, MCTN \cite{pham2019found} guarantees the model's robustness to the missing modality case by learning a joint representation through cyclic translation from the source modality to the target modality.
However, these methods suffer from the following limitations:
(\romannumeral1) inadequate interactions based on individual samples lack the mining of holistically structured semantics. (\romannumeral2) Failure to model cross-category correlations leads to loss of sentiment-relevant information and confusing distributions among categories. (\romannumeral3)  Coarse supervision ignores the semantic and distributional alignment.

To address the above issues, we present a \textbf{Corr}elation-decoupled \textbf{K}nowledge \textbf{D}istillation (CorrKD) framework for the MSA task under uncertain missing modalities. 
There are three core contributions in CorrKD based on the tailored components.
Specifically, 
(\romannumeral1) the proposed sample-level contrastive distillation mechanism captures the holistic cross-sample correlations and transfers valuable supervision signals via sample-level contrastive learning. (\romannumeral2) Meanwhile, we design a category-guided prototype distillation mechanism that leverages category prototypes to transfer intra- and inter-category feature variations, thus delivering sentiment-relevant information and learning robust joint multimodal representations. (\romannumeral3) Furthermore, we introduce a response-disentangled consistency distillation strategy to optimize sentiment decision boundaries and encourage distribution alignment by decoupling heterogeneous responses and maximizing mutual information between homogeneous sub-responses. Based on these components, CorrKD significantly improves MSA performance under uncertain missing-modality and complete-modality testing conditions on three multimodal benchmarks.

\vspace{-3pt}
\section{Related Work}
\subsection{Multimodal Sentiment Analysis}
MSA aims to understand and analyze human sentiment utilizing multiple modalities.  
Mainstream MSA studies \cite{hazarika2020misa,han2021improving, sun2022cubemlp,li2023decoupled,yang2024robust,yang2023context,yang2023target,yang2024SuCi,yang2024MCIS,yang2022contextual} focus on designing complex fusion paradigms and interaction mechanisms to enhance the performance of sentiment recognition. For instance, CubeMLP \cite{sun2022cubemlp} utilizes three independent multi-layer perceptron units for feature-mixing on three axes. 
However, these approaches based on complete modalities cannot be deployed in real-world applications.
Mainstream solutions for the missing modality problem can be summarized in two categories: (i) generative methods ~\cite{du2018semi, luo2023multimodal, lian2023gcnet, wang2023distribution} and (ii) joint learning methods~\cite{pham2019found, wang2020transmodality,zeng2022tag,liu2024modality}. 
Reconstruction methods generate missing features and semantics in modalities based on available modalities. For example, TFR-Net \cite{yuan2021transformer} leverages the feature reconstruction module to guide the extractor to reconstruct missing semantics. MVAE \cite{du2018semi} solves the modality missing problem by the semi-supervised multi-view deep generative framework. Joint learning efforts refer to learning joint multimodal representations utilizing correlations among modalities. For instance, MMIN \cite{zhao2021missing} generates robust joint multimodal representations via cross-modality imagination. TATE \cite{zeng2022tag} presents a tag encoding module to guide the network to focus on missing modalities. 
However, the aforementioned approaches fail to account for the correlations among samples and categories, leading to inadequate compensation for the missing semantics in modalities.
In contrast, we design effective learning paradigms to adequately capture potential inter-sample and inter-category correlations.

\subsection{Knowledge Distillation}
Knowledge distillation utilizes additional supervisory information from the pre-trained teacher's network to assist in the training of the student's network~\cite{hinton2015distilling}. 
Knowledge distillation methods can be roughly categorized into two types, distillation from intermediate features \cite{   kim2018paraphrasing, park2019relational,tian2019contrastive, yim2017gift} and responses~\cite{cho2019efficacy, furlanello2018born, mirzadeh2020improved, yang2019snapshot, zhao2017pyramid}.
Many studies \cite{hu2020knowledge, rahimpour2021cross, kumar2019online, wang2023learnable, xia2023robust} employ knowledge distillation for MSA tasks with missing modalities.
The core concept of these efforts is to transfer ``dark knowledge'' from teacher networks trained by complete modalities to student networks trained by missing modalities. 
The teacher model typically produces more valuable feature presentations than the student model.
For instance, \cite{hu2020knowledge} utilizes the complete-modality teacher network to implement supervision on the unimodal student network at both feature and response levels.
Despite promising outcomes, they are subject to several significant limitations:
(i) Knowledge transfer is limited to individual samples, overlooking the exploitation of clear correlations among samples and among categories.
(ii) Supervision on student networks is coarse-grained and inadequate, without considering the potential alignment of feature distributions. 
To this end, we propose a correlation-decoupled knowledge distillation framework that facilitates the learning of robust joint representations by refining and transferring the cross-sample, cross-category, and cross-target correlations.
\label{sec:related}

\begin{figure*}[t]
  \centering
  \includegraphics[width=\textwidth]{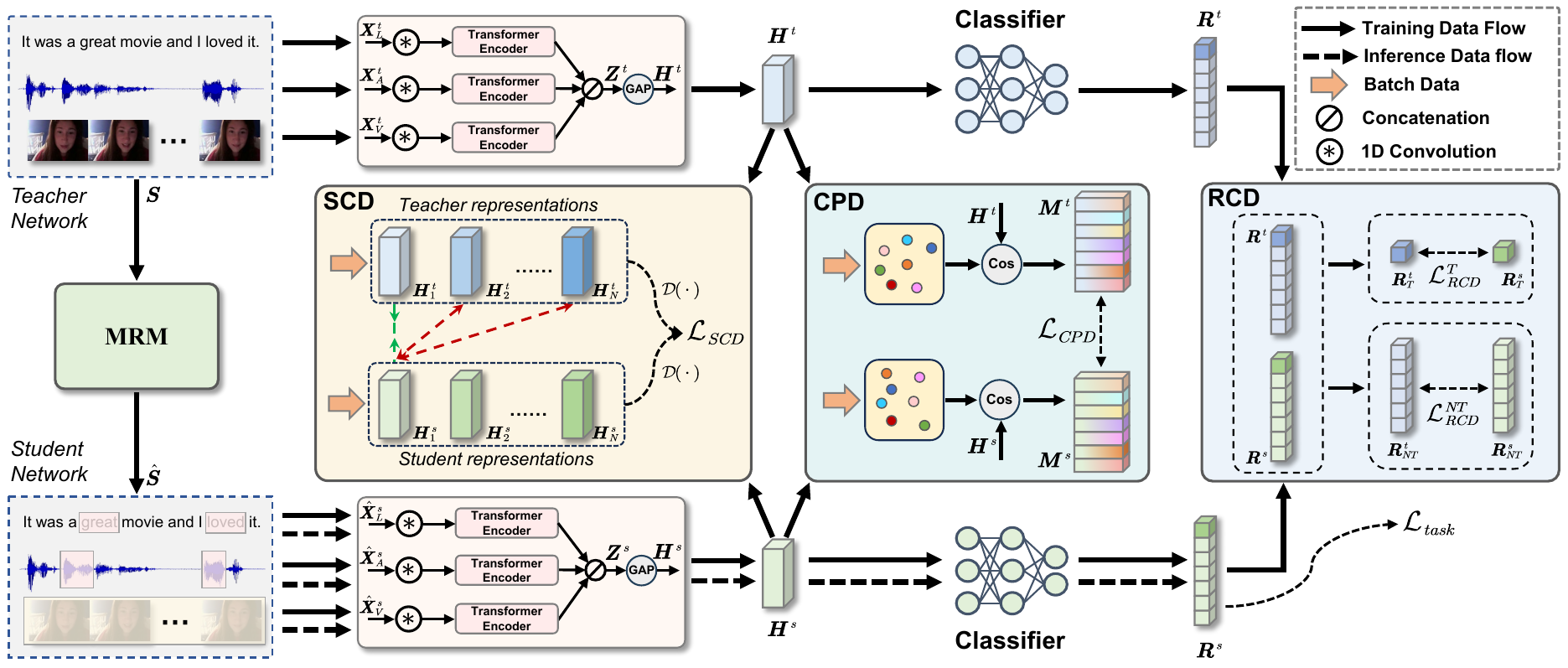}
  \caption{The structure of our CorrKD, which consists of three core components: Sample-level Contrastive Distillation (SCD) mechanism, Category-guided Prototype Distillation (CPD) mechanism, and Response-disentangled Consistency Distillation (RCD) strategy.
 }
  \label{fig:overall_framework1}
\end{figure*}

\section{Methodology}
\subsection{Problem Formulation}
Given a multimodal video segment with three modalities as $\bm{S}=[\bm{X}_L, \bm{X}_A, \bm{X}_V]$, where $\bm{X}_L\in \mathbb{R}^{T_L\times d_L}, \bm{X}_A \in \mathbb{R}^{T_A\times d_A}$, and  $\bm{X}_V\in \mathbb{R}^{T_V\times d_V}$ denote language, audio, and visual modalities, respectively. 
$T_{m}(\cdot)$ is the sequence length and $d_{m}(\cdot)$ is the embedding dimension, where $m \in \{L,A,V\}$.
Meanwhile, the incomplete modality is denoted as $\hat{\bm{X}}_m$. 
We define two missing modality cases to simulate the most natural and holistic challenges in real-world scenarios: 
(i) \textit{intra-modality missingness}, which indicates some frame-level features in the modality sequences are missing. (ii) \textit{inter-modality missingness}, which denotes some modalities are entirely missing.
Our goal is to recognize the utterance-level sentiments by utilizing the multimodal data with missing modalities.

\subsection{Overall Framework}
\Cref{fig:overall_framework1} illustrates the main workflow of CorrKD.
The teacher network and the student network adopt a consistent structure but have different parameters. 
During the training phase, our CorrKD procedure is as follows: 
(\romannumeral1) we train the teacher network with complete-modality samples and then freeze its parameters.
(\romannumeral2) Given a video segment sample $\bm{S}$, we generate a missing-modality sample 
$\hat{\bm{S}}$ with the Modality Random Missing (MRM) strategy. MRM simultaneously performs intra-modality missing and inter-modality missing, and the raw features of the missing portions are replaced with zero vectors. $\bm{S}$ and $\hat{\bm{S}}$ are fed into the initialized student network and the trained teacher network, respectively.
(\romannumeral3) We input the samples $\bm{S}$ and $\hat{\bm{S}}$ into the modality representation fusion module to obtain the joint multimodal representations $\bm{H}^t$ and $\bm{H}^s$.
(\romannumeral4) The sample-level contrastive distillation mechanism and the category-guided prototype distillation mechanism are utilized to learn the feature consistency of $\bm{H}^t$ and $\bm{H}^s$. 
(\romannumeral5) These representations are fed into the task-specific fully-connected layers and the softmax function to obtain the network responses $\bm{R}^t$ and $\bm{R}^s$.
(\romannumeral6) The response-disentangled consistency distillation strategy is applied to maintain consistency in the response distribution, and then $\bm{R}^s$ is used to perform classification.
In the inference phase, testing samples are only fed into the student network for downstream tasks.
Subsequent sections provide details of the proposed components.
\subsection{Modality Representation Fusion}
We introduce the extraction and fusion processes of modality representations using the student network as an example. The incomplete modality 
$\hat{\bm{X}}_m^s \in \mathbb{R}^{T_m \times d_m}$  with $m \in \{L,A,V\} $ is fed into the student network. 
Firstly, $\hat{\bm{X}}^s_m$ passes through a 1D temporal convolutional layer with kernel size $3 \times 3$ and adds the positional embedding \cite{vaswani2017attention} to obtain the preliminary representations, denoted as $\hat{\bm{F}}_m^s = \bm{W}_{3 \times 3}(\hat{\bm{X}}_m^s) + PE(T_m, d)\in\mathbb{R}^{T_m \times d}$.
Each $\bm{F}_m^s$ is fed into a Transformer~\cite{vaswani2017attention} encoder $\mathcal{F}_\phi^s(\cdot)$, capturing the modality dynamics of each sequence through the self-attention mechanism to yield representations $\bm{E}_m^s$, denoted as $\bm{E}_m^s = \mathcal{F}_\phi^s(\bm{F}_m^s)$.
The representations $\bm{E}_m^s$ are concatenated to obtain $\bm{Z}^s$, expressed as $\bm{Z^s} = [\bm{E}_L^s, \bm{E}_A^s, \bm{E}_V^s] \in \mathbb{R}^{T_m \times 3d}$. 
Subsequently, $\bm{Z}^s$ is fed into the Global Average Pooling (GAP) to further enhance and refine the features, yielding the joint multimodal representation $\bm{H}^s \in \mathbb{R}^{3d}$.
Similarly, the joint multimodal representation generated by the teacher network is represented as $\bm{H}^t \in \mathbb{R}^{3d}$.

\subsection{Sample-level Contrastive Distillation}
Most previous studies of MSA tasks with missing modalities \cite{rahimpour2021cross, xia2023robust, wang2023learnable} are sub-optimal, exploiting only one-sided information within a single sample and neglecting to consider comprehensive knowledge across samples.
To this end, we propose a Sample-level Contrastive Distillation (SCD) mechanism that enriches holistic knowledge encoding by implementing contrastive learning between sample-level representations of student and teacher networks.
This paradigm prompts models to sufficiently capture intra-sample dynamics and inter-sample correlations to generate and transfer valuable supervision signals, thus precisely recovering the missing semantics.
The rationale of SCD is to take contrastive learning within all mini-batches, constraining the representations in two networks originating from the same sample to be similar, and the representations originating from different samples to be distinct.

Specifically, given a mini-batch with $N$ samples $\bm{B} = \{\bm{S}_0, \bm{S}_1, \cdots, \bm{S}_N\}$, we obtain their sets of joint multimodal representations in teacher and student networks, denoted as $\{\bm{H}^w_1, \bm{H}^w_2, \cdots,  \bm{H}^w_N\}$ with $w \in \{t,s\}$.
For the same input sample, we narrow the distance between the joint representations of the teacher and student networks and enlarge the distance between the representations for different samples. 
The contrastive distillation loss is formulated as follows:
\begin{equation}
\small
    \mathcal{L}_{SCD} = \sum_{i=1}^N\sum_{j=1,j\neq i}^N\mathcal{D}(\bm{H}^s_i,\bm{H}^t_i)^2 + max\{0, \eta - \mathcal{D}(\bm{H}^s_i,\bm{H}^t_j)\}^2,
\end{equation}
where $\mathcal{D}(\bm{H}^s, \bm{H}^t) = \left\| \bm{H}^s  - \bm{H}^t \right\|_2  $ , $ \left\| \cdot \right\|_2$ represents $\ell_2$ norm function, and $\eta$ is the predefined distance boundary.
When negative pairs are distant enough (\emph{i.e.}, greater than boundary $\eta$), the loss is set to $0$, allowing the model to focus on other pairs.
Since the sample-level representation contains holistic emotion-related semantics, such a contrastive objective facilitates the student network to learn more valuable knowledge from the teacher network.

\subsection{Category-guided Prototype Distillation}
MSA data usually suffers from the dilemmas of high intra-category diversity and high inter-category similarity.
Previous approaches \cite{ hu2020knowledge, rahimpour2021cross,kumar2019online} based on knowledge distillation to address the modality missing problem simply constrain the feature consistency of the teacher and student networks.
The rough manner lacks consideration of cross-category correlation and feature variations, leading to ambiguous feature distributions.
To this end, we propose a Category-guided Prototype Distillation (CPD) mechanism, with the core insight of refining and transferring knowledge of intra- and inter-category feature variations via category prototypes,
which is widely utilized in the field of few-shot learning \cite{snell2017prototypical}. The category prototype represents the embedding center of every sentiment category, denoted as:
\begin{equation}
\small
    \bm{c}_k = \frac{1}{|\bm{B}_k|}\sum_{\bm{S}_i \in \bm{B}_k}{\bm{H}_i},
\end{equation}
where $\bm{B}_k$ denotes the set of samples labeled with category $k$ in the mini-batch, and $\bm{S}_i$ denotes the $i$-th sample in $\bm{B}_k$.
The intra- and inter-category feature variation of the sample $\bm{S}_i$ is defined as follows:
\begin{equation}
\small
    \bm{M}_k(i) = \frac{\bm{H}_i \, \bm{c}_k^\top}{\left\| \bm{H}_i \right\|_2  \left\| \bm{c}_k \right\|_2 },
\end{equation}
where $\bm{M}_k(i)$  denotes the similarity between the sample $\bm{S}_i$ and the prototype $\bm{c}_k$.
If the sample $\bm{S}_i$ is of category $k$, $\bm{M}_k(i)$ represents intra-category feature variation. Otherwise, it represents inter-category feature variation. 
The teacher and student networks compute similarity matrices $\bm{M}^t$ and $\bm{M}^s$, respectively. 
We minimize the squared Euclidean distance between the two similarity matrices to maintain the consistency of two multimodal representations.
The prototype distillation loss is formulated as:
\begin{equation}
\small
    \mathcal{L}_{CPD}=\frac{1}{N K} \sum_{i=1}^N \sum_{k=1}^K\left\|\bm{M}_k^s(i)-\bm{M}_k^t(i)\right\|_2,
\end{equation}
where $K$ is the category number of the mini-batch.

\subsection{Response-disentangled Consistency Distillation}
Most knowledge distillation studies~\cite{  kim2018paraphrasing, park2019relational,   tian2019contrastive, yim2017gift} focus on extracting knowledge from intermediate features of networks.
Although the model's response (\ie, the predicted probability of the model's output) presents a higher level of semantics than the intermediate features, response-based methods achieve significantly worse performance than feature-based methods \cite{wang2021knowledge}. 
Inspired by \cite{zhao2022decoupled}, the model's response consists of two parts: (i) Target Category Response (TCR), which represents the prediction of the target category and describes the difficulty of identifying each training sample. (ii) Non-Target Category Response (NTCR), which denotes the prediction of the non-target category and reflects the decision boundaries of the remaining categories to some extent.
The effects of TCR and NTCR in traditional knowledge distillation loss are coupled, \ie, high-confidence TCR leads to low-impact NTCR, thus inhibiting effective knowledge transfer.
Consequently, we disentangle the heterogeneous responses and constrain the consistency between the homogeneous responses.
From the perspective of information theory, knowledge consistency between responses can be characterized as maintaining high mutual information between teacher and student networks \cite{ahn2019variational}. 
This schema captures beneficial semantics and encourages distributional alignment.
\begin{figure*}[t]
  \centering
  \includegraphics[width=1.0\linewidth]{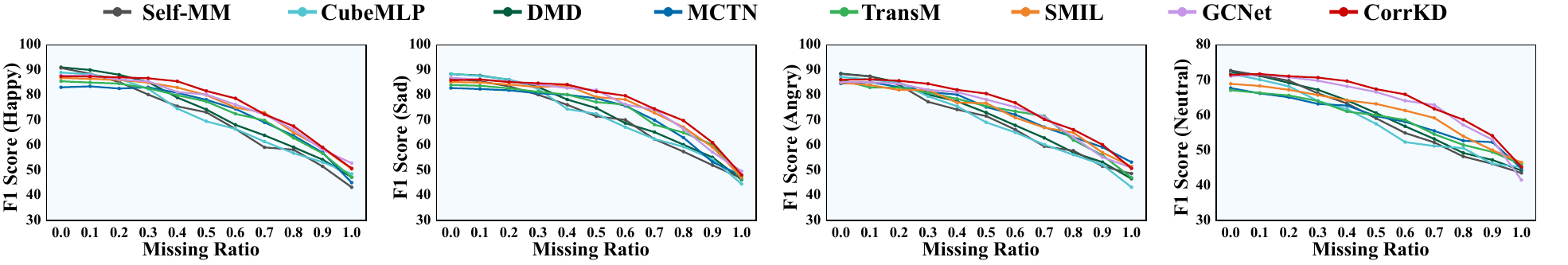}
  \caption{
  Comparison results of intra-modality missingness on IEMOCAP. We comprehensively report the F1 score for the happy, sad, angry, and neutral categories at various missing ratios.
 }
  \label{comp_intra_2}
\end{figure*}

\begin{figure}[t]
  \centering
  \includegraphics[width=1.0\linewidth]{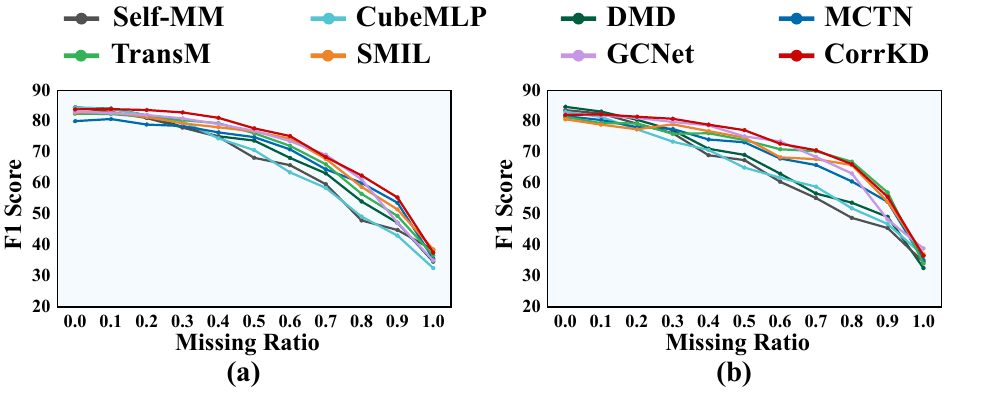}
  \vspace{-15pt}
  \caption{Comparison results of intra-modality missingness on (a) MOSI and (b) MOSEI. We report the F1 score at various ratios.
 }
  \label{comp_intra_1}
\end{figure}
Specifically, the joint multimodal representation $\bm{H}^w$  with $w\in\{t,s\}$ of teacher and student networks pass through fully-connected layers and softmax function to obtain response $\bm{R}^w$. Based on the target indexes, we decouple the response $\bm{R}^w$ to obtain TCR $\bm{R}^w_T$ and NTCR $\bm{R}^w_{NT}$.
Define $\bm{Q} \in \bm{\mathcal{Q}}$ and $\bm{U} \in \bm{\mathcal{U}}$ as two random variables.
Formulaically, the marginal probability density functors of $\bm{Q}$ and $\bm{U}$ are denoted as $P(\bm{Q})$ and $P(\bm{U})$. $P(\bm{Q},\bm{U})$ is regarded as the joint probability density functor. 
The mutual information between $\bm{Q}$ and $\bm{U}$ is represented as follows:
\begin{equation}
\small
    I(\bm{Q}, \bm{U})=\int_{\bm{\mathcal{Q}}} \int_{\bm{\mathcal{U}}} P(\bm{Q}, \bm{U}) \log \left(\frac{P(\bm{Q}, \bm{U})}{P(\bm{Q}) P(\bm{U})}\right) d \bm{Q} d \bm{U}.
\end{equation}
The mutual information $I(\bm{Q},\bm{U})$ can be written as the Kullback-Leibler divergence between the joint probability distribution $\mathbb{P}_{\bm{Q}\bm{U}}$ and the product of the marginal distributions $\mathbb{P}_{\bm{Q}} \mathbb{P}_{\bm{U}}$, denoted as $I(\bm{Q}, \bm{U})=\mathrm{D}_{K L}\left(\mathbb{P}_{\bm{Q} \bm{U}} \| \mathbb{P}_{\bm{Q}} \mathbb{P}_{\bm{U}}\right).$
For efficient and stable computation, the Jensen-Shannon divergence~\cite{hjelm2018learning} is employed in our case to estimate the mutual information, which is denoted as follows:
\begin{equation}
\small
\begin{aligned}
     I(\bm{Q}, \bm{U}) &\geq 
     \hat{I}_\theta^{(\mathrm{JSD})}(\bm{Q}, \bm{U}) \\ &=
      \mathbb{E}_{P(\bm{Q}, \bm{U})}\left[-\log \left(1+e^{-\mathcal{F}_\theta(\bm{Q}, \bm{U})}\right)\right] \\
    & -\mathbb{E}_{P(\bm{Q}) P(\bm{U})}\left[\log \left(1+e^{\mathcal{F}_\theta(\bm{Q}, \bm{U})}\right)\right],
\end{aligned}
\end{equation}
where $\mathcal{F}_\theta: \bm{Q} \times \bm{U} \rightarrow \mathbb{R}$
is formulated as an instantiated statistical network with parameters $\theta$.
We only need to maximize the mutual information without focusing on its precise value.
Consequently, the distillation loss based on the mutual information estimation is formatted as follows:
\begin{equation}
\small
    \mathcal{L}_{RCD} = \mathcal{L}_{RCD}^T + \mathcal{L}_{RCD}^{NT} = -I(\bm{R}^t_T,\bm{R}^s_T) - I(\bm{R}^t_{NT},\bm{R}^s_{NT}).
\end{equation}
Finally, the overall training objective $\mathcal{L}_{total}$ is expressed as $\mathcal{L}_{total} = \mathcal{L}_{task} + \mathcal{L}_{SCD} + \mathcal{L}_{CPD} + \mathcal{L}_{RCD}$, where $\mathcal{L}_{task}$ is the standard cross-entropy loss.

\section{Experiments}
\subsection{Datasets and Evaluation Metrics}
We conduct extensive experiments on three MSA datasets with word-aligned data, including MOSI \cite{zadeh2016mosi}, MOSEI \cite{zadeh2018multimodal}, and IEMOCAP \cite{busso2008iemocap}. 
\textbf{MOSI} is a realistic dataset that comprises 2,199 short monologue video clips.
There are 1,284, 229, and 686 video clips in train, valid, and test data, respectively. 
\textbf{MOSEI} is a dataset consisting of 22,856  video clips, which has 16,326, 1,871, and 4,659 samples in train, valid, and test data.
Each sample of MOSI and MOSEI is labeled by human annotators with a sentiment score of -3 (strongly negative) to +3 (strongly positive). On the MOSI and MOSEI datasets, we utilize weighted F1 score computed for positive/negative classification results as evaluation metrics.
\textbf{IEMOCAP} dataset consists of 4,453 samples of video clips. Its predetermined data partition has 2,717, 798, and 938 samples in train, valid, and test data.
As recommended by \cite{wang2019words}, four emotions (\emph{i.e.,} happy, sad, angry, and neutral) are selected for emotion recognition. 
For evaluation, we report the F1 score for each category.

\subsection{Implementation Details}
\paragraph{\textbf{Feature Extraction.}}
The Glove embedding~\cite{pennington2014glove} is used to convert the video transcripts to obtain a 300-dimensional vector for the language modality.
For the audio modality, we employ the COVAREP toolkit \cite{degottex2014covarep} to extract 74-dimensional acoustic features, including 12 Mel-frequency cepstral coefficients (MFCCs), voiced/unvoiced segmenting features, and glottal source parameters.
For the visual modality, we utilize the  Facet \cite{imotions2017facial} to indicate 35 facial action units, recording facial movement to express emotions.

\begin{table*}[t]
\centering
\caption{Comparison results under inter-modality missing and complete-modality testing conditions on MOSI and MOSEI.}
\renewcommand{\arraystretch}{1.2}
\setlength{\tabcolsep}{15pt}
\label{comp_inter_1}
\resizebox{\textwidth}{!}{%
\begin{tabular}{cccccccccc}
\toprule
\multirow{2}{*}{Dataset} & \multirow{2}{*}{Models} & \multicolumn{8}{c}{Testing Conditions}                                                                                                \\ \cmidrule{3-10} 
                         &                         & \{$l$\}          & \{$a$\}          & \{$v$\}          & \{$l,a$\}        & \{$l,v$\}        & \{$a,v$\}        & Avg.           & \{$l,a,v$\}      \\ \midrule
\multirow{8}{*}{MOSI}    & Self-MM \cite{yu2021learning}                 & 67.80          & 40.95          & 38.52          & 69.81          & 74.97          & 47.12          & 56.53          & \textbf{84.64} \\
                         & CubeMLP \cite{sun2022cubemlp}                 & 64.15          & 38.91          & 43.24          & 63.76          & 65.12          & 47.92          & 53.85          & 84.57          \\
                         & DMD \cite{li2023decoupled}                     & 68.97          & 43.33          & 42.26          & 70.51          & 68.45          & 50.47          & 57.33          & 84.50          \\
                         & MCTN \cite{pham2019found}                    & 75.21          & 59.25          & 58.57          & 77.81          & 74.82          & 64.21          & 68.31          & 80.12          \\
                         & TransM \cite{wang2020transmodality}                  & 77.64          & 63.57          & 56.48          & 82.07          & 80.90          & 67.24          & 71.32          & 82.57          \\
                         & SMIL \cite{ma2021smil}                    & 78.26          & \textbf{67.69}          & 59.67          & 79.82          & 79.15          & 71.24          & 72.64          & 82.85          \\
                         & GCNet \cite{lian2023gcnet}                   & 80.91          & 65.07          & 58.70          & \textbf{84.73} & \textbf{83.58} & 70.02          & 73.84          & 83.20          \\
                         & \textbf{CorrKD} & \textbf{81.20} & 66.52 & \textbf{60.72} & 83.56          & 82.41          & \textbf{73.74} & \textbf{74.69} & 83.94          \\ \midrule
\multirow{8}{*}{MOSEI}   & Self-MM \cite{yu2021learning}                 & 71.53          & 43.57          & 37.61          & 75.91          & 74.62          & 49.52          & 58.79          & 83.69          \\
                         & CubeMLP  \cite{sun2022cubemlp}                & 67.52          & 39.54          & 32.58          & 71.69          & 70.06          & 48.54          & 54.99          & 83.17          \\
                         & DMD \cite{li2023decoupled}                     & 70.26          & 46.18          & 39.84          & 74.78          & 72.45          & 52.70          & 59.37          & \textbf{84.78} \\
                         & MCTN  \cite{pham2019found}                   & 75.50          & 62.72          & 59.46          & 76.64          & 77.13          & 64.84          & 69.38          & 81.75          \\
                         & TransM \cite{wang2020transmodality}                  & 77.98          & 63.68          & 58.67          & 80.46          & 78.61          & 62.24          & 70.27          & 81.48          \\
                         & SMIL \cite{ma2021smil}                    & 76.57          & 65.96          & 60.57          & 77.68          & 76.24          & 66.87          & 70.65          & 80.74          \\
                         & GCNet \cite{lian2023gcnet}                   & 80.52          & \textbf{66.54} & 61.83          & \textbf{81.96} & 81.15          & 69.21          & 73.54          & 82.35          \\
                         & \textbf{CorrKD} & \textbf{80.76} & 66.09          & \textbf{62.30} & 81.74          & \textbf{81.28} & \textbf{71.92} & \textbf{74.02} & 82.16          \\ \bottomrule
\end{tabular}%
}
\end{table*}

\begin{table*}[t]
\caption{Comparison results under six testing conditions of inter-modality missingness and the complete-modality condition on IEMOCAP.}
\renewcommand{\arraystretch}{1.1}
\setlength{\tabcolsep}{18pt}
\label{comp_inter_2}
\resizebox{\textwidth}{!}{%
\begin{tabular}{cccccccccc}
\toprule
\multirow{2}{*}{Models}  & \multirow{2}{*}{Categories} & \multicolumn{8}{c}{Testing Conditions}                                                                                         \\ \cmidrule{3-10} 
                         &                          & \{$l$\}         & \{$a$\}         & \{$v$\}         & \{$l,a$\}       & \{$l,v$\}       & \{$a,v$\}       & Avg.          & \{$l,a,v$\}     \\ \midrule
\multirow{4}{*}{Self-MM \cite{yu2021learning}}                & Happy                    & 66.9          & 52.2          & 50.1          & 69.9          & 68.3          & 56.3          & 60.6          & 90.8          \\
                                        & Sad                      & 68.7          & 51.9          & 54.8          & 71.3          & 69.5          & 57.5          & 62.3          & 86.7          \\
                                        & Angry                    & 65.4          & 53.0          & 51.9          & 69.5          & 67.7          & 56.6          & 60.7          & 88.4          \\
                                        & Neutral                  & 55.8          & 48.2          & 50.4          & 58.1          & 56.5          & 52.8          & 53.6          & \textbf{72.7} \\ \midrule
\multirow{4}{*}{CubeMLP \cite{sun2022cubemlp}}                & Happy                    & 68.9          & 54.3          & 51.4          & 72.1          & 69.8          & 60.6          & 62.9          & 89.0          \\
                                        & Sad                      & 65.3          & 54.8          & 53.2          & 70.3          & 68.7          & 58.1          & 61.7          & \textbf{88.5} \\
                                        & Angry                    & 65.8          & 53.1          & 50.4          & 69.5          & 69.0          & 54.8          & 60.4          & 87.2          \\
                                        & Neutral                  & 53.5          & 50.8          & 48.7          & 57.3          & 54.5          & 51.8          & 52.8          & 71.8          \\ \midrule
\multirow{4}{*}{DMD \cite{li2023decoupled}}                    & Happy                    & 69.5          & 55.4          & 51.9          & 73.2          & 70.3          & 61.3          & 63.6          & \textbf{91.1} \\
                                        & Sad                      & 65.0          & 54.9          & 53.5          & 70.7          & 69.2          & 61.1          & 62.4          & 88.4          \\
                                        & Angry                    & 64.8          & 53.7          & 51.2          & 70.8          & 69.9          & 57.2          & 61.3          & \textbf{88.6} \\
                                        & Neutral                  & 54.0          & 51.2          & 48.0          & 56.9          & 55.6          & 53.4          & 53.2          & 72.2          \\ \midrule
\multirow{4}{*}{MCTN \cite{pham2019found}}                   & Happy                    & 76.9          & 63.4          & 60.8          & 79.6          & 77.6          & 66.9          & 70.9          & 83.1          \\
                                        & Sad                      & 76.7          & 64.4          & 60.4          & 78.9          & 77.1          & 68.6          & 71.0          & 82.8          \\
                                        & Angry                    & 77.1          & 61.0          & 56.7          & 81.6          & 80.4          & 58.9          & 69.3          & 84.6          \\
                                        & Neutral                  & 60.1          & 51.9          & 50.4          & 64.7          & 62.4          & 54.9          & 57.4          & 67.7          \\ \midrule
\multirow{4}{*}{TransM \cite{wang2020transmodality}}                 & Happy                    & 78.4          & 64.5          & 61.1          & 81.6          & 80.2          & 66.5          & 72.1          & 85.5          \\
                                        & Sad                      & 79.5          & 63.2          & 58.9          & 82.4          & 80.5          & 64.4          & 71.5          & 84.0          \\
                                        & Angry                    & 81.0          & 65.0          & 60.7          & \textbf{83.9}          & 81.7          & 66.9          & 73.2          & 86.1          \\
                                        & Neutral                  & 60.2          & 49.9          & 50.7          & 65.2          & 62.4          & 52.4          & 56.8          & 67.1          \\ \midrule
\multirow{4}{*}{SMIL \cite{ma2021smil}}                   & Happy                    & 80.5          & 66.5          & 63.8          & 83.1          & 81.8          & 68.2          & 74.0          & 86.8          \\
                                        & Sad                      & 78.9          & 65.2          & 62.2          & 82.4          & 79.6          & 68.2          & 72.8          & 85.2          \\
                                        & Angry                    & 79.6          & \textbf{67.2} & 61.8          & 83.1          & 82.0          & 67.8          & 73.6          & 84.9          \\
                                        & Neutral                  & 60.2          & 50.4          & 48.8          & 65.4          & 62.2          & 52.6          & 56.6          & 68.9          \\ \midrule
\multirow{4}{*}{GCNet \cite{lian2023gcnet}}                  & Happy                    & 81.9          & 67.3          & 66.6          & 83.7          & \textbf{82.5} & 69.8          & 75.3          & 87.7          \\
                                        & Sad                      & 80.5          & 69.4          & 66.1          & \textbf{83.8} & 81.9          & 70.4          & 75.4          & 86.9          \\
                                        & Angry                    & 80.1          & 66.2          & 64.2          & 82.5          & 81.6          & \textbf{68.1} & 73.8          & 85.2          \\
                                        & Neutral                  & 61.8          & 51.1          & 49.6          & 66.2          & 63.5          & 53.3          & 57.6          & 71.1          \\ \midrule
\multirow{4}{*}{\textbf{CorrKD}} & Happy                    & \textbf{82.6} & \textbf{69.6} & \textbf{68.0} & \textbf{84.1} & 82.0          & \textbf{70.0} & \textbf{76.1} & 87.5          \\
                                        & Sad                      & \textbf{82.7} & \textbf{71.3} & \textbf{67.6} & 83.4          & \textbf{82.2} & \textbf{72.5} & \textbf{76.6} & 85.9          \\
                                        & Angry                    & \textbf{82.2} & 67.0          & \textbf{65.8} & \textbf{83.9} & \textbf{82.8} & 67.3          & \textbf{74.8} & 86.1          \\
                                        & Neutral                  & \textbf{63.1} & \textbf{54.2} & \textbf{52.3} & \textbf{68.5} & \textbf{64.3} & \textbf{57.2} & \textbf{59.9} & 71.5          \\ \bottomrule
\end{tabular}%
}
\end{table*}

\noindent\textbf{Experimental Setup.} All models are built on the Pytorch \cite{paszke2017automatic} toolbox with NVIDIA Tesla V100 GPUs. 
The Adam optimizer \cite{kingma2014adam} is employed for network optimization.
For MOSI, MOSEI, and IEMOCAP, the detailed hyper-parameter settings are as follows: the learning rates are $\{4e-3, 2e-3, 4e-3\}$, the batch sizes are $\{64, 32, 64 \}$, the epoch numbers are $\{50, 20, 30\}$, the attention heads are $\{10, 8, 10\}$, and the distance boundaries $\eta$ are $\{1.2, 1.0, 1.4\}$. The embedding dimension is $40$ on all three datasets. The hyper-parameters are determined via the validation set.
The raw features at the modality missing positions are replaced by zero vectors.
To ensure an equitable comparison, we re-implement the state-of-the-art (SOTA) methods using the publicly available codebases and combine them with our experimental paradigms.
All experimental results are averaged over multiple experiments using five different random seeds.

\subsection{Comparison with State-of-the-art Methods}
We compare CorrKD with seven representative and reproducible SOTA methods, including complete-modality methods: Self-MM \cite{yu2021learning}, CubeMLP \cite{sun2022cubemlp}, and DMD \cite{li2023decoupled}, and missing-modality methods: 1) joint learning methods (\emph{i.e.}, MCTN \cite{pham2019found} and TransM \cite{wang2020transmodality}), and 2) generative methods (\emph{i.e.}, SMIL \cite{ma2021smil} and GCNet \cite{lian2023gcnet}). 
Extensive experiments are implemented to thoroughly evaluate the robustness and effectiveness of CorrKD in the cases of intra-modality and inter-modality missingness.

\noindent \textbf{Robustness to Intra-modality Missingness.}
We randomly drop frame-level features in modality sequences with ratio $p \in \{0.1, 0.2, \cdots, 1.0\}$ to simulate testing conditions of intra-modality missingness.
Figures~\ref{comp_intra_2} and \ref{comp_intra_1} show the performance curves of models with various $p$ values, which intuitively reflect the model's robustness. 
We have the following important observations.
(\romannumeral1) As the ratio $p$ increases, the performance of all models decreases. This phenomenon demonstrates that intra-modality missingness leads to a considerable loss of sentiment semantics and fragile joint multimodal representations.
(\romannumeral2) Compared to the complete-modality methods (\ie, Self-MM, CubeMLP, and DMD), our CorrKD achieves significant performance advantages in the missing-modality testing conditions and competitive performance in the complete-modality testing conditions. The reason is that complete-modality methods are based on the assumption of data completeness, whereas customized training paradigms for missing modalities perform better at capturing and reconstructing valuable sentiment semantics from incomplete multimodal data.
(\romannumeral3) Compared to the missing-modality methods, our CorrKD exhibits the strongest robustness. Benefiting from the decoupling and modeling of inter-sample, inter-category, and inter-response correlations by the proposed correlation decoupling schema, the student network acquires informative knowledge to reconstruct valuable missing semantics and produces robust multimodal representations.

\noindent \textbf{Robustness to Inter-modality Missingness.}
In \Cref{comp_inter_1} and ~\ref{comp_inter_2}, we drop some entire modalities in the samples to simulate testing conditions of inter-modality missingness.
The notation ``$\{l\}$'' indicates that only the language modality is available, while audio and visual modalities are missing. ``$\{l, a, v\}$'' represents the complete-modality testing condition where all modalities are available. ``Avg.'' indicates the average performance across six missing-modality testing conditions.
We present the following significant insights.
\begin{figure}[t]
  \centering
  \includegraphics[width= \linewidth]{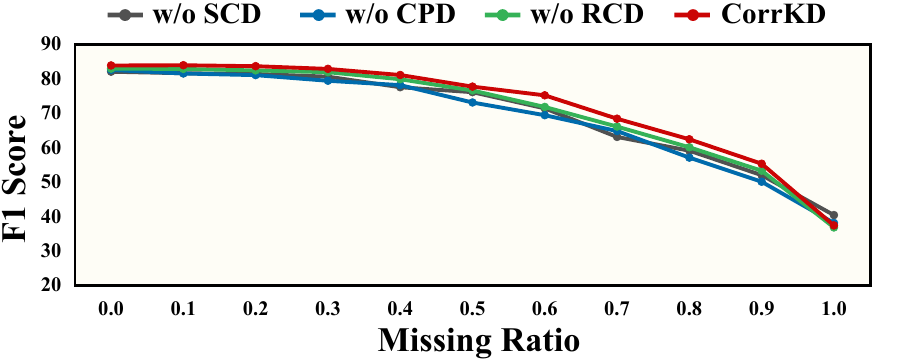}
  \caption{ Ablation results of intra-modality missingness using various missing ratios on MOSI. 
 }
  \label{abla_intra_mosi}
\end{figure}
\begin{table}[t]
\centering
\caption{Ablation results for the testing conditions of inter-modality missingness on MOSI.}
\label{ablation_inter_mosi}
\renewcommand{\arraystretch}{1.3}
\setlength{\tabcolsep}{4pt}
\resizebox{\columnwidth}{!}{%
\begin{tabular}{ccccccccc}
\toprule
\multirow{2}{*}{Models} & \multicolumn{8}{c}{Testing Conditions}                                                                             \\ \cmidrule{2-9} 
                        & \{$l$\}          & \{$a$\}          & \{$v$\}          & \{$l,a$\} & \{$l,v$\} & \{$a,v$\}        & Avg.           & \{$l,a,v$\} \\ \midrule
\textbf{CorrKD} & \textbf{81.20} & \textbf{66.52} & \textbf{60.72} & \textbf{83.56} & \textbf{82.41} & \textbf{73.74} & \textbf{74.69} & \textbf{83.94} \\
w/o SCD                  & 78.80          & 64.96          & 57.49          & 81.95          & 80.53          & 71.05          & 72.46          & 82.13          \\
w/o CPD                  & 79.23          & 63.72          & 57.83          & 80.11          & 79.45          & 70.53          & 71.81          & 82.67          \\
w/o RCD                  & 79.73          & 65.32          & 59.21          & 82.14          & 81.05          & 72.18          & 73.27          & 83.05          \\ \bottomrule
\end{tabular}%
}
\end{table}
\begin{figure*}[t]
  \centering
  \includegraphics[width=\linewidth]{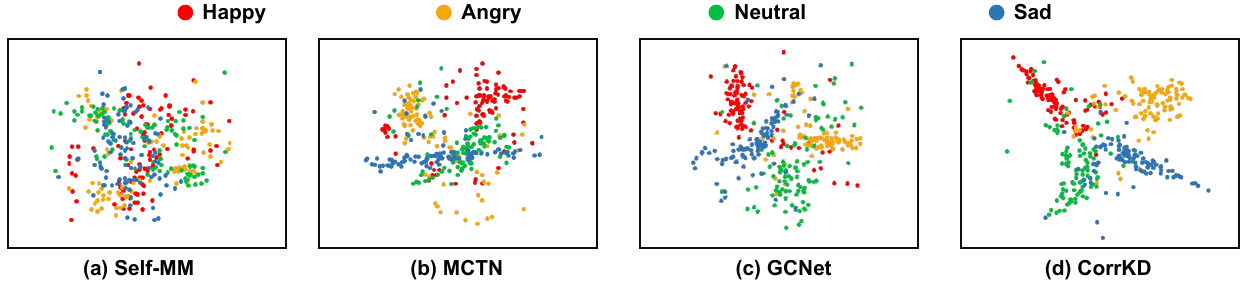}
  \caption{
 Visualization of representations from different methods with four emotion categories on the IEMOCAP testing set.
The default testing conditions contain intra-modality missingness (\ie, missing ratio $p = 0.5$ ) and inter-modality missingness (\ie, only the language modality is available). 
The \textcolor{red}{red}, \textcolor{orange}{orange}, \textcolor{green}{green}, and \textcolor{blue}{blue} markers represent the happy, angry, neutral, and sad emotions, respectively.
 }
  \label{vis3}
\end{figure*}
(\romannumeral1) Inter-modality missingness causes performance degradation for all models, suggesting that the integration of complementary information from heterogeneous modalities enhances the sentiment semantics within joint representations.
(\romannumeral2) In the testing conditions of the inter-modality missingness, our CorrKD has superior performance among the majority of metrics, proving its strong robustness.  
For example, on the MOSI dataset, CorrKD's average F1 score is improved by $0.85\%$ compared to GCNet, and in particular by $3.72\%$ in the testing condition where language modality is missing (\ie, $\{a,v\}$).
The merit stems from the proposed framework's capability of decoupling and modeling potential correlations at multiple levels to capture discriminative and holistic sentiment semantics.
(\romannumeral3)  In the unimodal testing conditions, the performance of CorrKD with only the language modality favorably outperforms other cases, with comparable results to the complete-modality case. In the bimodal testing conditions, cases containing the language modality perform the best, even surpassing the complete-modality case in individual metrics. This phenomenon proves that language modality encompasses the richest knowledge information and dominates the sentiment inference and missing semantic reconstruction.

\subsection{Ablation Studies}
To validate the effectiveness and necessity of the proposed mechanisms and strategies in CorrKD, we conduct ablation studies under two missing-modality cases on the MOSI dataset, as shown in \Cref{ablation_inter_mosi} and \Cref{abla_intra_mosi}.
The principal findings are outlined as follows.
(\romannumeral1) 
When SCD is eliminated, there is a noticeable degradation in model performance under both missing cases.
This phenomenon suggests that mining and transferring comprehensive cross-sample correlations is essential for recovering missing semantics in student networks.
(\romannumeral2) 
The worse results under the two missing modality scenarios without CPD indicate that capturing cross-category feature variations and correlations facilitates deep alignment of feature distributions between both networks to produce robust joint multimodal representations.
(\romannumeral3) Moreover, we substitute the KL divergence loss for the proposed RCD. The declining performance gains imply that decoupling heterogeneous responses and maximizing mutual information between homogeneous responses motivate the student network to adequately reconstruct meaningful sentiment semantics.

\subsection{Qualitative Analysis}
To intuitively show the robustness of the proposed framework against modality missingness, we randomly choose 100 samples from each emotion category on the IEMOCAP testing set for visualization analysis.
The comparison models include Self-MM~\cite{yu2021learning} (\ie, complete-modality method), MCTN~\cite{pham2019found} (\ie, joint learning-based missing-modality method), and GCNet~\cite{lian2023gcnet} (\ie, generative-based missing-modality method). 
(i) As shown in \Cref{vis3}, Self-MM cannot address the modality missing challenge, as the representations of different emotion categories are heavily confounded, leading to the least favorable outcomes.
(ii) Although MCTN and GCNet somewhat alleviate the issue of indistinct emotion semantics, their effectiveness remains limited since the distribution boundaries of the different emotion representations are generally ambiguous and coupled.
(iii) Conversely, our CorrKD ensures that representations of the same emotion category form compact clusters, while representations of different categories are clearly separated. These observations confirm the robustness and superiority of our framework, as it sufficiently decouples inter-sample, inter-category and inter-response correlations.

\section{Conclusions}
In this paper,  we present a correlation-decoupled knowledge distillation framework (CorrKD) to address diverse missing modality dilemmas in the MSA task.
Concretely, we propose a sample-level contrast distillation mechanism that utilizes contrastive learning to capture and transfer cross-sample correlations to precisely reconstruct missing semantics.
Additionally, we present a category-guided prototype distillation mechanism that learns cross-category correlations through category prototypes, refining sentiment-relevant semantics for improved joint representations.
Eventually, a response-disentangled consistency distillation is proposed to encourage distribution alignment between teacher and student networks. 
Extensive experiments confirm the effectiveness of our framework.

\section*{Acknowledgements}
This work is supported in part by the Shanghai Municipal Science and Technology Committee of Shanghai Outstanding Academic Leaders Plan (No.\,21XD1430300), and in part by the National Key R\&D Program of China (No.\,2021ZD0113503).

{
    \small
    \bibliographystyle{ieeenat_fullname}
    \bibliography{main}
}

\end{document}